# Practical Design and Benchmarking of Generative AI Applications for Surgical Billing and Coding

John C. Rollman, MSTAT [a], Bruce Rogers, PhD [a], Hamed Zaribafzadeh, MS [a], Daniel Buckland, MD, PhD [b], Ursula Rogers, BS [a], Jennifer Gagnon, RN [a], Ozanan Meireles, MD [a], Lindsay Jennings, BPS [c], Jim Bennett, BS [a], Jennifer Nicholson, MEd [c], Nandan Lad, MD, PhD [d], Linda Cendales, MD [a], Andreas Seas, BS [d e f], Alessandro Martinino, MD [f], E. Shelley Hwang, MD, MPH [a], Allan D. Kirk, MD, PhD [a]

[a] Department of Surgery, Duke University Medical Center, Durham, North Carolina
[b] Department of Emergency Medicine, Duke University Medical Center, Durham, North Carolina
[c] Patient Revenue Management Organization, Duke University Health System, Durham, North Carolina
[d] Department of Neurosurgery, Duke University Medical Center, Durham, North Carolina
[e] Department of Biomedical Engineering, Duke Pratt School of Engineering, Durham, North Carolina
[f] Duke University School of Medicine, Durham, North Carolina

**Corresponding Author:**
John Rollman
Email: john.rollman@duke.edu

## ABSTRACT

**Background**

Healthcare has many manual processes that can benefit from automation and augmentation with Generative Artificial Intelligence (AI), the medical billing and coding process. However, current foundational Large Language Models (LLMs) perform poorly when tasked with generating accurate International Classification of Diseases, 10th edition, Clinical Modification (ICD-10-CM) and Current Procedural Terminology (CPT) codes. Additionally, there are many security and financial challenges in the application of generative AI to healthcare. We present a strategy for developing generative AI tools in healthcare, specifically for medical billing and coding, that balances accuracy, accessibility, and patient privacy.

**Methods**

We fine tune the PHI-3 Mini and PHI-3 Medium LLMs using institutional data and compare the results against the PHI-3 base model, a PHI-3 RAG application, and GPT-4o. We use the post operative surgical report as input and the patients billing claim the associated ICD-10, CPT, and Modifier codes as the target result. Performance is measured by accuracy of code generation, proportion of invalid codes, and the fidelity of the billing claim format.

**Results**

Both fine-tuned models performed better or as well as GPT-4o. The Phi-3 Medium fine-tuned model showed the best performance (ICD-10 Recall and Precision: 72%, 72%; CPT Recall and Precision: 77%, 79%; Modifier Recall and Precision: 63%, 64%). The Phi-3 Medium fine-tuned model only fabricated 1% of ICD-10 codes and 0.6% of CPT codes generated.

**Conclusions**

Our study shows that a small model that is fine-tuned on domain-specific data for specific tasks using a simple set of open-source tools and minimal technological and monetary requirements performs as well as the larger contemporary consumer models.

## BACKGROUND

With Generative AI interest on the rise, many in healthcare have begun exploring potential applications of the technology to address specific use cases to augment manual processes.[1,2] Many of these approaches center on reducing administrative burden on clinicians. These tools could result in reduced rates of physician burnout, and increased time spent with patients.[3,4] However, the healthcare industry, with its specific vocabulary, complex processes, and strict regulatory guardrails makes the application of generative AI in this domain a major challenge.[5–8]

Although foundational models, like GPT 4, have been shown to perform well on medical education questions, such as those found on the United States Medical Licensing Exam[9,10], they achieve mediocre results when interpreting and synthesizing the electronic health record (EHR) for technical applications like billing and coding.[11,12] Clinical language models, foundational LLMs trained on EHR and other clinical documents, have been developed to adapt language models to the nuances of healthcare.[13,14] Health systems have attempted to train their own foundational models; however, these models require significant resources to gather, clean, and prepare extensive amounts of data. They also require substantial technical infrastructure on the scale of hundreds of GPUs to train successfully.[13] An alternative to training a completely new model is to add domain specific knowledge and task expertise to an existing foundational model through fine-tuning, using full parameter supervised fine-tuning (SFT) or Parameter Efficient Fine-tuning (PEFT), and in-context learning, by way of prompt engineering or Retrieval Augmented Generation (RAG) systems.[15–17] Both fine-tuning and in-context learning substantially reduce the resource requirements making domain-specific generative AI more accessible.[18–20] The choice of one design over the other is not a clear-cut decision, and the resulting performance can depend on the task at hand.[19,21,22]

We sought to determine whether generative AI, specifically Large Language Models (LLMs), could be tailored to the surgical billing and coding process. Our primary goal was to augment and existing LLM to be performant at generating diagnostic and procedural codes for a given operative procedure. We sought to achieve this without restrictive resource requirements and with feasible integration into current coder workflows. Our work focuses on the practical development and application of generative AI in healthcare and documents the capabilities of adapting custom LLMs to health system operations. Our methods prioritize entirely local development on limited technical infrastructure to ensure security and minimize barriers to reproducibility. We evaluate multiple design strategies for achieving this goal using real patient records from a mixed community/academic health system. As a result, we show that our strategies can produce results on par or better than the largest State of the Art (SOTA) models

## METHODS
### Data
We extracted the operative reports and associated billing codes for all outpatient and ambulatory surgical encounters from January 2017 through December 2022 performed across the health system. The health system includes an academic tertiary care center, two affiliated community hospitals, and two ambulatory surgery centers. The EHR data were queried from our enterprise data warehouse. The data were split into training, validation, and test sets using 60%, 20%, and 20% of the data respectively. The split was balanced across surgical encounter date to account for billing and coding practice changes.

**Data Flow and Security**

The primary input to the proposed system was the operative report that was generated by the operating surgeon, and the output was a billing claim generated by an LLM that consisted of the relevant ICD-10 diagnosis codes, the performed CPT codes for each provider, and the accompanying CPT code modifiers. To satisfy security and compliance requirements, the work was done in our Protected Analytics Computing Environment (PACE), a virtual machine (VM) with restricted network access and no data egress ability. This would provide the necessary precautions to allow the use of Protected Health Information (PHI) in the medical record data. The use of identifiable data was a priority, as it has been shown to be a key factor in model performance.[23] The VM was equipped with four NVIDIA a5000 Graphics Processing Units (GPUs) each with 24 gigabytes of video random access memory (VRAM).

**Language Model Configurations**

The experiment consisted of four design configurations of the Phi-3 family of models from Microsoft, primarily the Phi3-Mini.[24,25] The models were selected for their size, long context length, and robust performance.[24] The first configuration used the base Phi3-Mini model alone to establish a baseline efficacy. The second configuration combined the base model with a knowledge database into a RAG system to evaluate the efficacy of in-context learning. The third configuration implemented the Phi3-Mini model fine-tuned on institution-specific operative and billing data to evaluate the efficacy of fine-tuning. The fourth configuration employed the larger parameter model version, Phi-3-Medium fine-tuned on the same institution-specific data to evaluate the efficacy gains due to model size. The model weights were downloaded locally from HuggingFace and manually transferred to our protected environment. Lastly, GPT-4o was used to establish a comparative benchmark representing large state of the art (SOTA) models. All Phi-3 configurations utilized the 128k context-length model variant.

**RAG System Design**

The Facebook AI Similarity Search (FAISS) suite of tools was used to create an in-memory vector database.[26,27] OP notes from the training set were embedded using the all-MiniLM-L6-v2 model from the Sentence Transformers library.[28] The embedding model was chosen for its balance of embedding speed and overall performance for query-vector searches. To minimize system complexity, vector embeddings were stored using a flat architecture and utilized an exhaustive L2 search. The associated ICD-10s, CPTs, and modifiers were added to each embedded documents metadata. The RAG system used the input OP Note from the test set as the query and returned the billing code metadata from the top two most similar OP Notes in the vector database. This was done to match similar procedures rather than match OP Notes to exact ICD-10 and CPT descriptors. The resulting code examples were then appended to the model prompt as context.

**Fine-tuning**

Models were fine-tuned on the operative report – billing claim pairs from the training set. The training prompt was formatted using the following instruct template used during the model's pretraining.

```
<s><|system|>
You are an expert on medical coding and procedural billing
in the United States. Your task is to assist in creating an
appropriate billing claim given the provided operative
report. Every claim should include ICD-10-CM codes, CPT
codes from the American Medical Association, and the
modifiers for each CPT code.<|end|>

<|user|>
What ICD-10-CM diagnosis codes, CPT codes, and CPT
modifiers could be added to the billing claim for the
following procedure?<|placeholder1|>

Operative Report:
{operative note text}<|placeholder1|><|end|>

<|assistant|>
<|placeholder2|>ICD-10-CM Diagnoses:

{ICD-10-CM Codes}<|placeholder3|>

<|placeholder4|>CPT Codes with Modifiers:

Provider Name: {provider name}
Provider Billables:
CPT {#}: {CPT code} | Modifiers: {modifiers for CPT code #}
 | Description: {description of CPT code #}<|placeholder5|>
```

We utilized the unused special tokens from the model's vocabulary to segment the prompt into logical components, with the aim of cueing the model to better respond to the specific task.

We performed SFT using Quantized Low Rank Adapters (QLoRA), a highly efficient and performant PEFT method. The method was chosen to accommodate the limited technical resources available, reduce training time, and maximize future integration feasibility. PEFT methods target and re-parameterize only a fractional subset of model's total parameters, thus greatly reducing the training time and technical requirements. Since adapters are trained and stored separately from the base model, we can more efficiently store the model. The use of LoRA also allows us to swap out different trained adapters for different tasks for the same base model. We set the training hyperparameters according to the original QLoRA study to maximize performance and reduce catastrophic forgetting in the base model.[29] We used a rank of 64, alpha of 16, and dropout rate of 0.1. All the linear layer parameters of the model, 'o_proj',

'qkv_proj', 'gate_up_proj', 'down_proj', and 'lm_head' were targeted during SFT. The model weights were loaded into GPU memory in nf4 type 4-bit double quantization but saved in bfloat16. Computation was done in bfloat16. This meant that model weights were converted back and forth between 4-bit and bfloat16 during the SFT process. This reduced the overall memory footprint and resource requirements but added some additional processing overhead resulting in an increase in training time. Flash Attention 2 was used to further optimize the model's computation speeds and reduce training time.[30]

Fine-tuning was done across 4 NVIDIA a5000 24GB GPUs on a single machine. The process was parallelized using DeepSpeed Zero Redundancy Optimizer (ZeRO) Stage-3.[31] DeepSpeed is a library that facilitates data, pipeline, and tensor parallelism together for highly efficient distributed training.[31–33] The Stage-3 configuration automatically shards the training optimizer states, gradients, and model parameters across data parallel workers, where a worker is a single GPU.

**Inference**
Inference for all 4 model configurations followed a standardized structure. All models were loaded into GPU memory with 4-bit quantization and Flash Attention 2 was utilized. Token generation was done in bfloat16 precision. To reduce inference latency, we merged the trained adapters from the fine-tuning back into their respective models. During inference, the model was replicated across the 4 GPUs and the evaluation test set was split across the 4 GPUs and processed in parallel using the PyTorch and Accelerate libraries. The input prompts were formatted for the whole test set prior to inference. The prompts for the fine-tuned configurations were structured as:

```
<s><|system|>
You are an expert on medical coding and procedural billing
in the United States. Your task is to assist in creating an
appropriate billing claim given the provided operative
report. Every claim should include ICD-10-CM codes, CPT
codes from the American Medical Association, and the
modifiers for each CPT code.<|end|>

<|user|>
What ICD-10-CM diagnosis codes, CPT codes, and CPT
modifiers could be added to the billing claim for the
following procedure?<|placeholder1|>

Operative Report:
{operative note text}<|placeholder1|><|end|>

<|assistant|>
```

```
        <|placeholder2|>
```

For the RAG system, the insertion of the relevant examples was done during prompt formatting, prior to inference, to save time on evaluation. The resulting prompt was:

```
        <s><|system|>
        You are an expert on medical coding and procedural billing
        in the United States. Your task is to assist in creating an
        appropriate billing claim given the provided operative
        report. Every claim should include ICD-10-CM codes, CPT
        codes from the American Medical Association, and the
        modifiers for each CPT code.<|end|>
        <|user|>
        What ICD-10-CM diagnosis codes, CPT codes, and CPT
        modifiers could be added to the billing claim for the
        following procedure?<|placeholder1|>

        Operative Report:

        {operative note text}<|placeholder1|>

        The following are examples from similar procedures:

        Relevant Example Claim 1:

        {example billing claim from similar OP Note}

        Relevant Example Claim 2:

        {example billing claim from similar OP Note}

        Provide the answer in the following format:

        ICD-10-CM Diagnoses:
        XXX.XXX, XXX.XXX, ...

        CPT Codes with Modifiers:
        CPT Code 1: ##### | Modifiers: XX, XX, ...
        CPT Code 2: ##### | Modifiers: XX, XX, ...
        <|end|>
        <|assistant|>
```

An additional formatting instruction and example were added to the base configuration and RAG configuration prompts to ensure consistent output format. The base configuration prompt structure was as follows:

```
<s><|system|>
You are an expert on medical coding and procedural billing
in the United States. Your task is to assist in creating an
appropriate billing claim given the provided operative
report. Every claim should include ICD-10-CM codes, CPT
codes from the American Medical Association, and the
modifiers for each CPT code.<|end|>
<|user|>
What ICD-10-CM diagnosis codes, CPT codes, and CPT
modifiers could be added to the billing claim for the
following procedure?<|placeholder1|>

{operative note text}<|placeholder1|>

Provide the answer without descriptions in the following
format:

ICD-10-CM Diagnoses:
XXX.XXX, XXX.XXX, ...

CPT Codes with Modifiers:
CPT Code 1: ##### | Modifiers: XX, XX, ... | Description:
...
CPT Code 2: ##### | Modifiers: XX, XX, ... | Description:
...
<|end|>
```

The HuggingFace Transformers library was used to create the inference pipeline. Prompt-responses were processed one at a time due to varying input tokenization lengths. The generation settings were set to a maximum of 512 new tokens with a repetition penalty of 1.1. For the generation method, we chose to pursue a conservative and deterministic strategy utilizing a greedy search with no sampling and 1 beam. Our goal was to limit response variability, spurious output formats, and hallucinations. Greedy search, as a generation method, has been shown to provide robust comparative performance for similar models while retaining high throughput speeds.[34]

**Evaluation**

To assess the quality of the generated output from the inference process, we compared the generated claim ICD-10s, CPTs, and modifiers against the code on the true billing claim submitted for payment. The primary measures of success were the accuracy rate and validity of exact code matches and consistency of output format. While other studies have measured whether a code was in the same family or semantically close to the actual code,[11,35] "being close" was not helpful to the medical coder's and billing staff workflows.[36] Additionally, a lack of consistent output format from the proposed tool would complicate extracting the relevant codes, either by program or by the coder's themselves resulting in minimal value-add.

Codes were extracted from the output using regular expressions. If the codes were unable to be extracted due to formatting discrepancies, they were considered wrong. To quantify the extent of hallucinations, where the model produces a code that does not exist, we compared each generated code to a reference list of ICD-10s and CPTs from the appropriate cohort year. Both the proportion of fabricated codes and its complement, the proportion of valid codes, were calculated and reported in the results table. Recall and precision were calculated at the individual case level. The metrics were then averaged and F1 was calculated to assess the aggregate balanced performance.

We define precision as:
$$\text{Mean Precision} = \frac{\sum_c \frac{|M_c \cap P_c|}{|M_c|}}{|C|}$$

and recall as:
$$\text{Mean Recall} = \frac{\sum_c \frac{|M_c \cap P_c|}{|P_c|}}{|C|}$$

where:
- Mc = set of codes generated by the model M for case c
- Pc = set of codes generated by the medical coder P for case c
- C = set of cases in cohort

F1 score was calculated as:
$$F = 2 * \frac{(Precision * Recall)}{(Precision + Recall)}$$

To quantify the robustness of output format consistency, we chose to evaluate the generated claims using the Recall-Oriented Understudy for Gisting Evaluation (ROUGE) metrics, primarily focusing on the ROUGE-L calculation, and the Metric for Evaluation of Translation with Explicit Ordering (METEOR). Both ROUGE and METEOR are used to evaluate textual similarity using word-based matching and provide a proximate measure of output structure. ROUGE-L identifies the longest common subsequence between the model output and the ground truth. The ROUGE-L Sum score calculates the sum of all the common subsequences between the output and ground truth.[37] METEOR utilizes exact, stem, and synonym word matching while penalizing incorrect word order.[38] METEOR assesses both the recall and precision of the matches. These methods offered enough flexibility for the varying billed code set sizes and configurations.

**RESULTS**
**Data Extraction and Processing**
There were a total of 192,585 surgical encounters extracted across multiple surgical services, as shown in Table 1. The operative notes were tokenized for exploratory analysis. Figure 1 shows the distribution of token lengths and the corresponding summary statistics for the operative notes used as input for fine tuning and inference. Figure 2 shows the frequency of ICD-10 codes and CPT codes, respectively.

*Table 1*

| Total Encounters | | 192585 |
|---|---|---|
| Patient Age, mean (SD) | | 54.7 (21.9) |
| Sex, n (%) | F | 109068 (56.6) |
| | M | 83517 (43.4) |
| Year of Surgery, n (%) | 2017 | 27539 (14.3) |
| | 2018 | 29805 (15.5) |
| | 2019 | 31489 (16.4) |
| | 2020 | 29006 (15.1) |
| | 2021 | 36444 (18.9) |
| | 2022 | 38302 (19.9) |
| Service, n (%) | Anes/ Pain Mgmt | 1735 (0.9) |
| | Cardiothoracic | 1581 (0.8) |
| | Dermatology | 461 (0.2) |
| | Gastroenterology | 415 (0.2) |
| | General Surgery | 25677 (13.3) |
| | Gynecology | 10531 (5.5) |
| | Neurosurgery | 4872 (2.5) |
| | Obstetrics | 1761 (0.9) |
| | Ophthalmology | 59655 (31.0) |
| | Orthopedics | 46178 (24.0) |
| | Otolaryngology Head and Neck | 13744 (7.1) |
| | Pediatric Bone Marrow Transplant | 13 (0.0) |
| | Pediatric Dental Surgery | 52 (0.0) |
| | Pediatric Gastroenterology | 18 (0.0) |
| | Pediatric Hematology-Oncology | 6 (0.0) |
| | Pediatric Nephrology | 18 (0.0) |
| | Pediatric Neurology | 1 (0.0) |
| | Pediatric Rheumatology | 53 (0.0) |
| | Pediatric Surgery | 2331 (1.2) |
| | Plastic Surgery | 8228 (4.3) |
| | Podiatry | 1 (0.0) |
| | Pulmonary | 228 (0.1) |
| | Urology | 15026 (7.8) |

*Case counts broken out by patient age, sex, surgery year, and surgical service. Surgical service is defined by internal records and as listed on patient's medical record.*

*Figure 1*

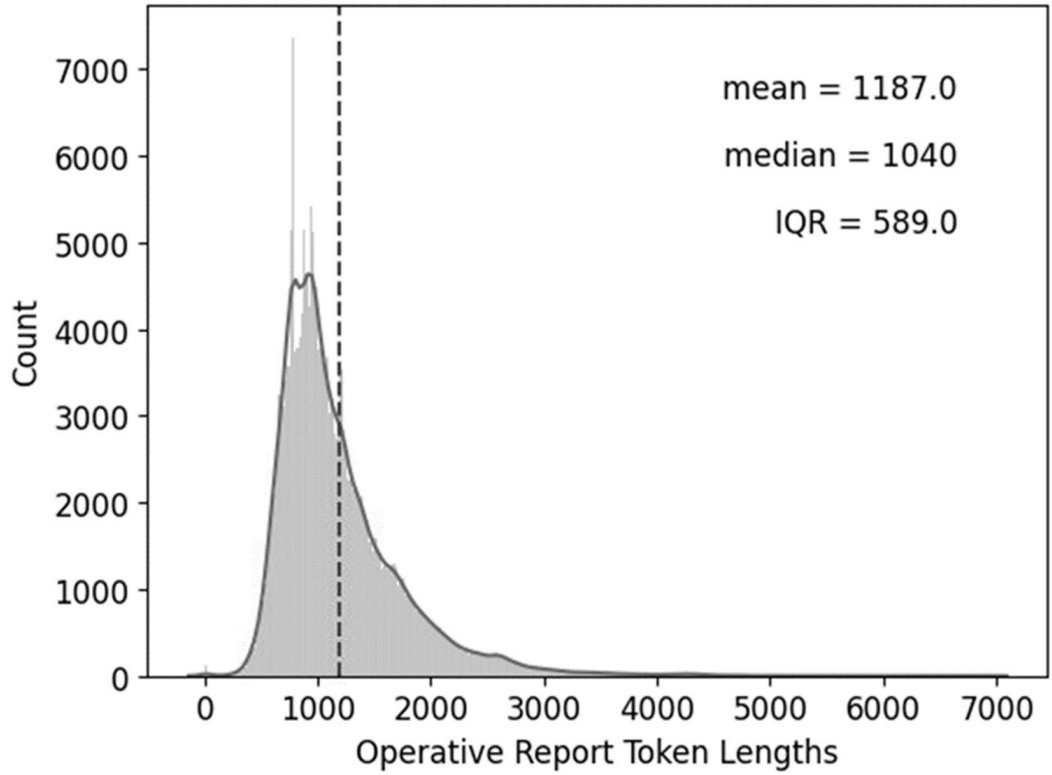

*Histogram of operative note token lengths. Dashed line indicates the mean token length across the distribution. Token length calculated using the Phi-3 Mini Model's tokenizer.*

*Figure 2*

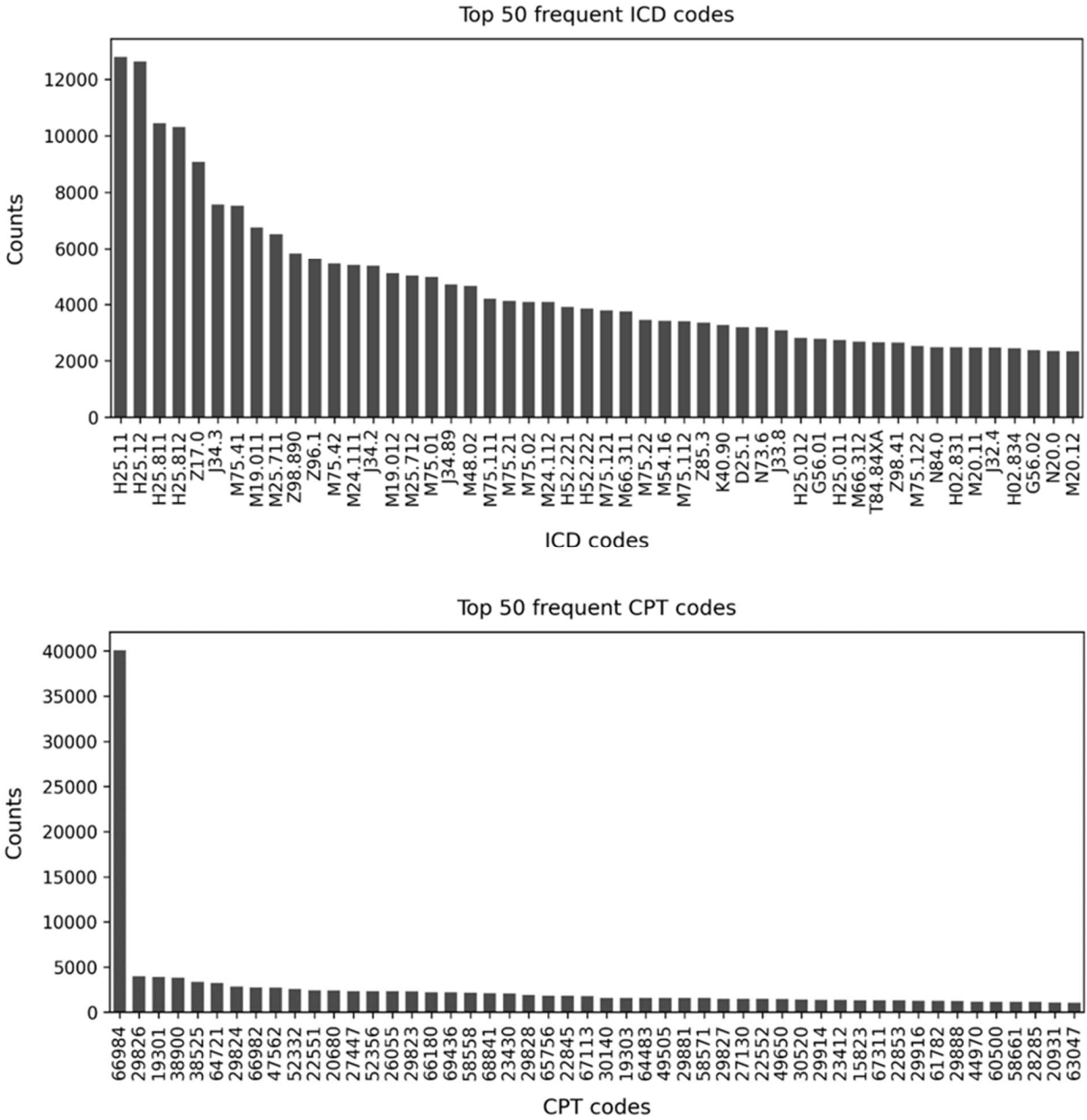

*Counts of ICD-10 and CPT codes across case population.*

**Fine Tuning**

The Phi-3-Mini took 7.3 hours to complete, and the total number of trained parameters was 201,612,288, about 5% of the total 3.8 billion tunable parameters. The Phi-3-Medium took 30.5 hours to complete, and the total number of trained parameters was 389,784,576, about 2.7% of the total 14 billion tunable parameters.

**Code Set Generation Analysis**

The results for exact code set match across ICD-10s, CPTs, and modifiers are shown in Table 2. Amongst the customized models, the Phi-3 Medium fine-tuned model performed the best across all metrics. Recall was scored at 65% for ICD-10s, 77% for CPTs, and 63% for modifiers. Precision was scored at 72% for ICD-10s, 79% for CPTs, and 64% for modifiers. This was followed by the Phi-3 Mini fine-tuned model, and then the Phi-3 Mini Base RAG. Both fine-tuned models performed better or as well as GPT-4o. The base Phi-3 variant performed the worst across all metrics and struggled with appending the correct modifiers to the claim. The base Phi-3 variant was also the only model to perform worse on CPT matching than ICD-10 matching. We calculate confidence intervals for the evaluation metrics by bootstrapping, shown in Figure 3. The validation set is sampled with replacement (N = 39052), and the mean precision and recall are calculated for the sample. This process is repeated for 1000 bootstrap iterations to estimate the distributions of the metrics.

*Table 2*

| Claim Component | Metric | Phi-3-mini | Phi-3-mini RAG | Fine-tuned Phi-3-mini | Fine-tuned Phi-3-medium | GPT-4o |
|---|---|---|---|---|---|---|
| ICD-10-CM | Full Match % | 5.2% | 23.5% | 40.0% | 43.7% | 30.7% |
| | Valid % | 50.3% | 76.9% | 97.9% | 99% | 96.6% |
| | Fabricated % | 49.7% | 23.1% | 2.1% | 1% | 3% |
| | Recall | 0.39 | 0.52 | 0.60 | 0.65 | 0.52 |
| | Precision | 0.33 | 0.48 | 0.68 | 0.72 | 0.58 |
| | F1 | 0.36 | 0.50 | 0.64 | 0.68 | 0.55 |
| CPT | Full Match % | 3.1% | 32.4% | 53.5% | 63.7% | 36.7% |
| | Valid % | 61.4% | 70.2% | 98.1% | 99.4% | 99.0% |
| | Fabricated % | 38.6% | 29.8% | 1.9% | 0.6% | 1.0% |
| | Recall | 0.30 | 0.63 | 0.69 | 0.77 | 0.72 |
| | Precision | 0.17 | 0.56 | 0.70 | 0.79 | 0.60 |
| | F1 | 0.22 | 0.59 | 0.69 | 0.78 | 0.66 |
| Modifier | Full Match % | 0.1% | 13.6% | 43.9% | 49.8% | 14.2% |
| | Recall | 0.01 | 0.32 | 0.61 | 0.63 | 0.30 |
| | Precision | 0.01 | 0.28 | 0.59 | 0.64 | 0.31 |
| | F1 | 0.01 | 0.30 | 0.60 | 0.64 | 0.30 |
| Structure | ROUGE L | 38.0 | 39.4 | 80.9 | 86.6 | 62.5 |
| | ROUGE L Sum | 39.7 | 40.6 | 81.7 | 87.4 | 64.7 |
| | METEOR Score | 0.42 | 0.34 | 0.68 | 0.87 | 0.62 |

*ICD-10-CM, CPT, Modifier, and Structure performance metrics for all model configurations.*

*Figure 3*

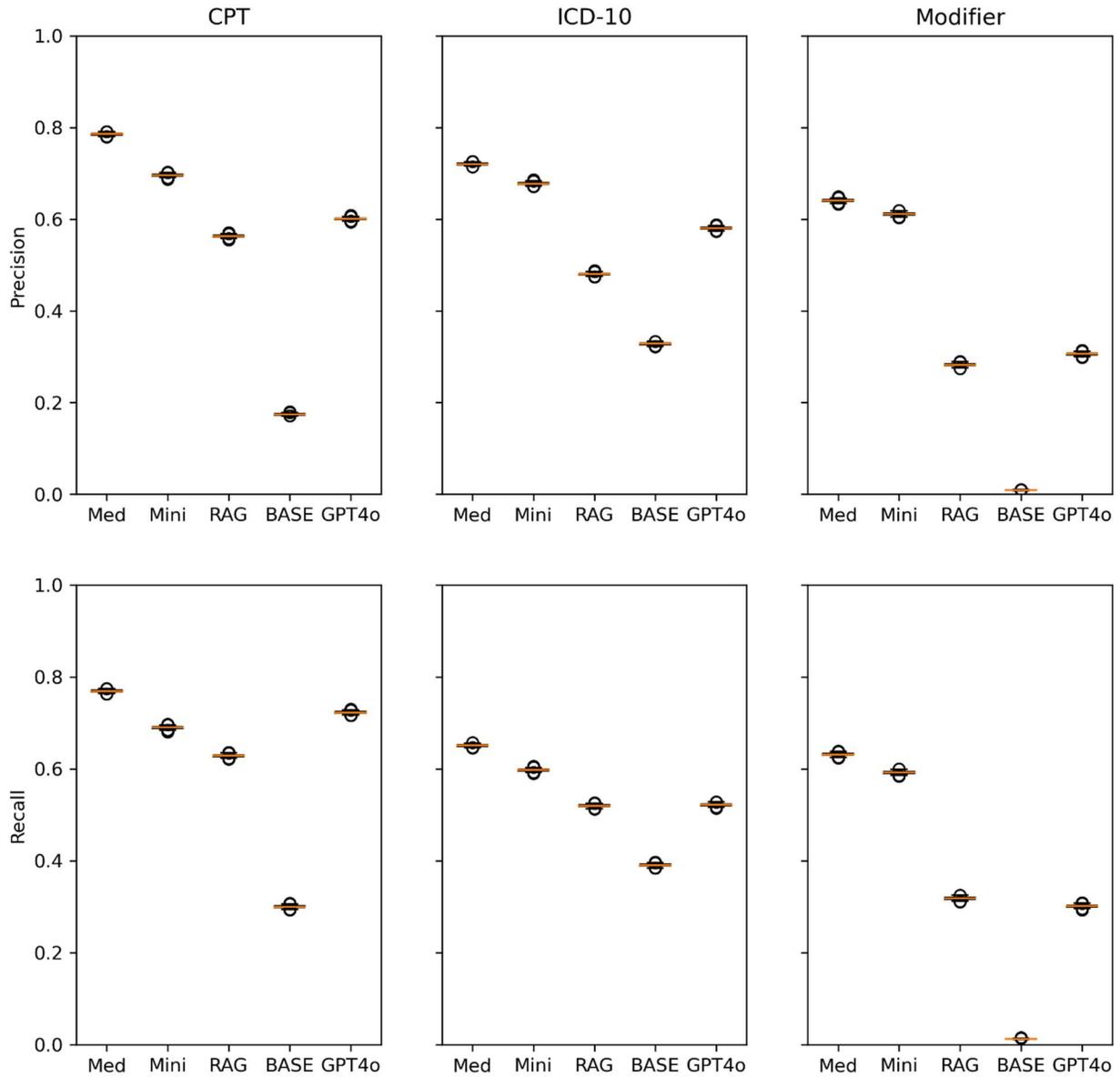

Box plots of bootstrapped model recall and precision for all configurations. Data were sampled with replacement (N = 39052) for 1000 bootstrap iterations. Outliers are included.

**Code Validity Analysis**

The percentage of valid codes and the corresponding hallucination rate, percent fabricated, are provided in Table X. The Phi-3 Medium fine-tuned model had the fewest occurrences of fabricated codes with only 1% of ICD-10 codes and 0.6% of CPT codes being fabricated. The Phi-3 Mini fine-tuned model had fewer fabricated ICD-10 codes than GPT-4o, 2.1% versus 3%, but a higher occurrence of CPT fabrication, 1.9% versus 1%. The base Phi-3 Mini and RAG configuration performed considerable worse than the others, with Phi-3 Mini base model fabricating ICD-10 codes nearly 50% of the time and CPT codes 39% of the time. The Phi-3 Mini RAG fabricated ICD-10 codes 23% of the time and CPT codes about 30% of the time.

**Format Consistency Analysis**
We assessed the consistency and robustness of the generated text output by comparing it against the ground truth data in the expected format. The Phi-3 Medium model performed the best with a ROUGE L score of 86.6 and a METEOR score of 0.87, followed by the fine-tuned mini variant, the RAG variant, and then the base variant. GPT- 4o outperformed the base and RAG variants but scored worse than the fine-tuned variants. The performance difference between models was consistent across the three scores. However, the Phi-3 Mini fine-tuned variant had a comparatively worse METEOR score than the Phi-3 Medium fine-tuned variant despite having similar ROUGE L Scores. This was likely due to the strictness of word ordering METEOR imposes over ROUGE. The disparity could be associated with the CPT descriptions in the generated claim, where the medium model produced descriptions that more closely matched the true AMA CPT Description, whereas the Mini model took liberties in semantics.

**DISCUSSION**
**Analysis**
Our study evaluated the efficacy of LLMs to generate post-surgical billing claim codes across 4 local application configurations. We tested two well-known methods for adding domain knowledge to LLMs, fine-tuning and in-context learning, compared them to the unaltered base model, and documented the resulting performance. More importantly, the work was accomplished utilizing realistic infrastructure and technical resources, serving as a model to drive AI with relevant health system integration under realistic limitations and model constraints.[39]

The Phi-3 Medium fine-tuned model performed the best out of all configurations. This was expected due to its larger parameter size. The altered variants all did moderately well at producing viable CPT and ICD-10 code sets without overproducing bad guesses. We find this to be a valuable result given that it reduces the checking a medical coder would have to do. Further, the balance between recall and precision is ideal given that missing a code completely is as much a detractor as producing an incorrect one. Given that medical coders balance case throughput with case accuracy for hundreds of encounters a day, such a tool could considerably streamline this process.

We found both fine-tuned models performed moderately well with CPT and ICD-10 codes given the only input information was an operative report. Medical coders generally have access to the entire patient record. This additional context is important for correctly assigning codes, especially for ICD-10 diagnosis codes. We expected to see a worse performance across the board for generating ICD-10 over CPT codes. When investigating the medical coder's workflow, it was noted that the diagnosis codes often come from information contained in the History and Physical (H&P) text in the patient record. Thus, incorporating this additional information would be key to improving model performance in generation of diagnosis codes.

Overall, the Phi-3 Mini fine-tuned model and RAG demonstrated strong resource-to-performance results, matching the performance of highly sophisticated flagship models such as GPT-4o. While direct comparison is difficult,[40,41] our models show competitive results to common industry tools.[36,42] This performance is particularly notable, given that these tools typically have access to the entire patient record and encompasses a wide range of Medical Coding background material. There is even some evidence showing comparative performance to medical coders.[43] When compared to the measured internal recall of our current Computer Assisted Coding (CAC) software (61%), both fine-tuned models match or slightly beat performance.

Importantly, both configurations were constructed with economic and integration viability as a priority. Their straightforward design and low resource requirements make reproducibility more accessible across institutions. They utilized entirely open-source tools and required minimal data pipeline changes to the current EHR data extraction process. Fine-tuning and inference were performed across 4 mid-sized GPUs but could have been less at the cost of increased time and a reduction in dataset size, although we note that fine-tuning does not require a large dataset.[29,44,45]

The configurations are good candidates for near-term integration and use in the current workflow for billing and coding. All the model configurations can feasibly be loaded on consumer hardware with the settings described in this paper. Additionally, there are more optimizations for deploying and serving this technology. Strategies such as paged attention, continuous batching, and caching provide fast inference throughput, minimal latency, and memory efficiency.[46] As an example, the Phi-3 Mini in its quantized state is only 1.7GB and can be loaded and run on most commercial cell phones. The cost to performance of these configurations makes them competitive against even the largest general models, such as GPT-4. In fact, GPT 4 reportedly struggled to produce viable and accurate codes when given the exact descriptors.[11,47]

Despite our demonstrated successes, we must return to the core question at hand, did this technology improve the delivery of healthcare? While there are certain indirect benefits to the coders workflow, none of the LLM formulations tested with real clinical notes performed at a level of competence that would lead to reduced FTE or training requirements of coders/billers in our clinical system, nor would the performance result in improvement in a time-to-submitted-for-claims metric. At best, these methods could provide a cost-efficient alternative to expensive industry coding software suites. In their current state LLMs are not a replacement for medical coders but can be a powerful tool when applied strategically and appropriately to streamline this labor-intensive process.

**Limitations and Future Work**

As discussed, we prioritized an economical and accessible methodology for tapping into the potential of generative AI for healthcare. Across the entire application, there are many configuration opportunities and variables that could be tweaked to further improve the performance. We note this as a primary limitation of our study and, more broadly, a limitation of studying AI in healthcare. We plan to continue to benchmark different configurations as well as incorporate the H&P in future iterations.

**CONCLUSION**

Our study shows that a small model that is fine-tuned on domain-specific data for specific tasks using a simple set of open-source tools performs as well as the largest SOTA general models for a fraction of the resource requirements. Moreover, we have demonstrated that by aiming for simple and economical solutions we can make considerable progress in the adoption of AI in healthcare workflows.